\pdfoutput=1

\documentclass[11pt]{article}

\usepackage{EMNLP2023}

\usepackage{times}
\usepackage{latexsym}

\usepackage[T1]{fontenc}

\usepackage[utf8]{inputenc}

\usepackage{microtype}

\usepackage{inconsolata}

\usepackage{ascii}
\usepackage{hyperref}
\usepackage{amsmath}
\usepackage{amsfonts}
\usepackage{algorithm}
\usepackage[noend]{algpseudocode}
\usepackage{bm}
\usepackage{bbm}
\usepackage{booktabs}
\usepackage{balance}
\usepackage{color,colortbl}
\usepackage{CJKutf8}
\usepackage{graphicx} 
\usepackage{graphics}
\usepackage{multirow}
\usepackage{makecell}
\usepackage{paralist}
\usepackage{subfig}
\usepackage{textcomp}
\usepackage{threeparttable}
\usepackage[normalem]{ulem}
\usepackage{arydshln}
\usepackage{enumitem}

\usepackage{cleveref}
\crefformat{section}{\S#2#1#3} 

\definecolor{jred}{RGB}{225, 11, 11}
\definecolor{jblue}{RGB}{41, 52, 190}
\definecolor{jgreen}{RGB}{18, 141, 21}
\definecolor{jorange}{RGB}{255, 127, 80}
\definecolor{jgray}{RGB}{10, 10, 10}
\definecolor{silver}{RGB}{192, 192, 192}

\definecolor{wxjiao}{RGB}{18, 141, 21}

\title{\texttt{ParroT}: Translating during Chat using Large Language Models tuned with Human Translation and Feedback
}

\author{
Wenxiang Jiao$^1$\thanks{~~Corresponding author.}
\enspace Jen-tse Huang$^{1,2}$ \enspace Wenxuan Wang$^{1,2}$ \enspace Zhiwei He$^{1,3}$ \enspace Tian Liang$^{1,4}$ \\ 
\bf Xing Wang$^1$ \enspace Shuming Shi$^1$ \enspace Zhaopeng Tu$^1$ \\[1ex]
$^1$Tencent AI Lab \enspace $^2$The Chinese University of Hong Kong \\
$^3$Shanghai Jiao Tong University \enspace $^4$Tsinghua Shenzhen International Graduate School 
\\
{\asciifamily \normalsize \tt \{joelwxjiao,brightxwang,shumingshi,zptu\}@tencent.com} \\
{\asciifamily \normalsize \tt \{jthuang,wxwang\}@cse.cuhk.edu.hk
} \\
{\asciifamily \normalsize \tt zwhe.cs@sjtu.edu.cn \enspace liangt21@mails.tsinghua.edu.cn 
} \\[1ex]
}

\begin{document}
\maketitle
\begin{abstract}
Large language models~(LLMs) like ChatGPT have exhibited remarkable abilities on a wide range of natural language processing~(NLP) tasks, including various machine translation abilities accomplished during chat. 
However, these models are only accessible through restricted APIs, which creates barriers to new research and advancements in the field.
Therefore, we propose ParroT, a framework to enhance and regulate the translation abilities during chat based on open-source LLMs~(e.g., LLaMA), human-written translation and feedback data.
Specifically, ParroT reformulates translation data into the instruction-following style, and introduces a “Hint” field for incorporating extra requirements to regulate the translation process. Accordingly, we propose three instruction types for finetuning ParroT models, including translation instruction, contrastive instruction, and error-guided instruction.
Experiments on Flores subsets and WMT22 test sets suggest that translation instruction improves the translation performance of vanilla LLMs significantly while error-guided instruction can lead to further improvement, which demonstrates the importance of learning from low-quality translations annotated by humans.
We also demonstrate the potential of automatic evaluation tools in providing quality information of translations, when constructing error-guided instructions for directions that lack human annotation data.
Please refer to our Github project for more implementation details: \url{https://github.com/wxjiao/ParroT}.
\end{abstract}

\section{Introduction}
\label{sec:introduction}

Large language models (LLMs), designed in the instruction-following format, such as ChatGPT and GPT-4~\cite{openai2023gpt4}, have garnered considerable interest due to their remarkable abilities in comprehending instructions and generating human-like responses. These versatile models can efficiently perform a wide range of natural language processing (NLP) tasks within a single architecture, including question answering~\cite{Omar2023ChatGPTVT}, text summarization~\cite{Yang202ChatGPT4Summary}, grammatical error correction~\cite{wu2023chatgpt4gec}, and machine translation~\cite{jiao2023ischatgpt}. Consequently, they represent a significant stride toward the realization of artificial general intelligence (AGI).

\begin{figure}[t]
    \centering
    \includegraphics[width=1.0\columnwidth]{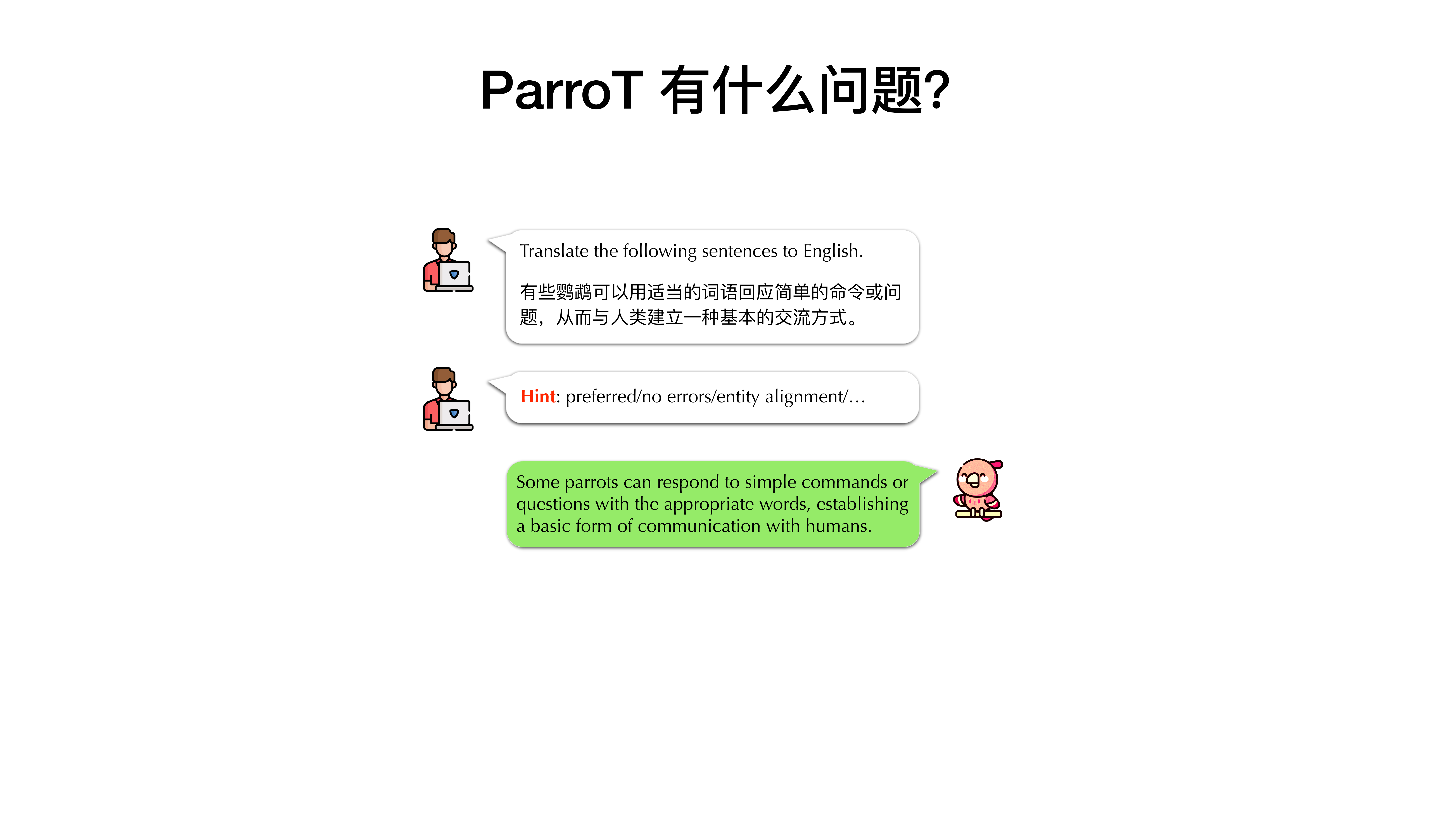}
    \caption{Framework of ParroT. Hints are (optional) extra requirements to regulate the translation process.}
    \label{fig:parrot}
\end{figure}

Machine translation, a quintessential NLP task, faces both challenges and opportunities presented by the emergence of LLMs. Traditional machine translation encompasses several sub-tasks~\cite{farhad2021wmt}, such as bilingual translation~\cite{vaswani2017attention}, multilingual translation~\cite{johnson2017google,jiao2022tencent}, terminology translation~\cite{wang2022template,hou2022adapters}, quality estimation~\cite{rei2020comet}, and automatic post-editing~\cite{pal2016neural}, among others. These tasks are typically addressed by individual models with limited cross-task interaction. However, current LLMs have the potential to revolutionize this inefficient approach and redefine the machine translation paradigm.
On one hand, LLMs can leverage the benefits of various sub-tasks and seamlessly transition between them using only natural language instructions. For instance, if a user is dissatisfied with a translation result, they can request the LLM to refine the translation implicitly (i.e., through automatic post-editing) or explicitly, by imposing constraints on specific entities (i.e., terminology translation).
On the other hand, LLMs are expected to enhance the explainability of machine translation, ultimately leading to further improvements in translation quality. For example, users may want LLMs to compare two translations of a sentence (i.e., quality estimation) and provide an explanation for the discrepancies (i.e., error analysis), which can then be addressed in a targeted manner by the LLM itself.
However, superior LLMs like ChatGPT and GPT-4 are only accessible through restricted APIs, which creates barriers to new research and advancements in the field.
Therefore, developing comprehensive machine translation abilities upon open-source LLMs has become a critical and challenging research problem.

In this paper, we propose the ParroT framework to enhance and regulate the translation abilities of LLMs during chat by leveraging existing human-written translation and feedback data.
To be compatible with chat, our framework reformulates translation data into the instruction-following style~\cite{alpaca}, and introduces a ``Hint'' field for incorporating extra requirements to guide the translation process.
Accordingly, we propose three distinct instruction types: (1) \textbf{Translation Instruction}, that asks LLMs to generate translations based on source sentences.
(2) \textbf{Contrastive Instruction}, that asks LLMs to generate the translations of two different systems with the preferred one at first. 
(3) \textbf{Error-Guided Instruction}, that asks LLMs to generate the translations with human-annotated errors as the hint.
The first instruction guarantees the basic translation ability of LLMs while the latter two regulate the LLMs to align with human feedbacks~\cite{ouyang2022InstructGPT,liu2023CoH}.
We adopt the open-source LLaMA~\cite{touvron2023llama} and BLOOM~\cite{scao2022bloom} models, and conduct instruction tuning on previous WMT validation data and Multidimensional Quality Metric~(MQM) human evaluation data. The resulting ParroT models are evaluated on Flores subsets and WMT22 test sets. 

Our main findings are summarized as below:
\begin{itemize}
    \item Translation instruction, as expected, can improve the translation performance of LLMs significantly, especially for directions from English to other languages.
    \item Error-guided instruction can further improve the performance when asking ParroT to generate translations with no error, indicating the importance of learning from low-quality translations annotated by humans.
    \item Parameter efficient finetuning with low-rank adaptation~\cite[LoRA,][]{hu2022lora} can prevent LLMs from overfitting, which achieves better performance on dominant languages but slows down the learning from other languages.
    \item We demonstrate the potential of automatic evaluation tools (i.e., COMET) in providing quality information of translations, when constructing error-guided instructions for directions that lack human annotation data.
\end{itemize}

\begin{table*}[t!]
\fontsize{9}{11}\selectfont
\centering
\caption{Instruction types for finetuning ParroT models.}
\resizebox{1.0\textwidth}{!}{
\begin{tabular}{l p{14cm}}
\toprule
\bf Type & \multicolumn{1}{c}{\bf Instruction} \\
\midrule
& Below is an instruction that describes a task, paired with an input that provides further context. Write a response that appropriately completes the request.{\color{gray}\textbackslash n\textbackslash n} \\
\midrule
Translation  & \colorbox{gray!20}{\#\#\# Instruction:}{\color{gray}\textbackslash n}Translate the following sentences from Chinese to English.{\color{gray}\textbackslash n\textbackslash n} \\[1ex]
& \colorbox{gray!20}{\#\#\# Input:}{\color{gray}\textbackslash n}\begin{CJK}{UTF8}{gkai}
检查情况显示，市场销售的粮油、肉类、水果、蔬菜、蛋奶等生活必需品供应充足，商品价格基本稳定，未发现严重违法违规行为，市场经营秩序总体平稳。\end{CJK}{\color{gray}\textbackslash n\textbackslash n} \\[0.3ex]
& \colorbox{green!20}{\color{jgreen}\#\#\# Response:}The inspection results showed that there was an adequate supply of daily necessities, including grain, oil, meat, fruit, vegetable, milk, and eggs in the market and commodity prices basically remain stable, the administration found no serious offensive and noncompliant conducts, and the market order remains stable on the whole. \\
\midrule
Contrastive & \colorbox{gray!20}{\#\#\# Instruction:}{\color{gray}\textbackslash n}Translate the following sentences from Chinese to English.{\color{gray}\textbackslash n\textbackslash n} \\[1ex] 
& \colorbox{gray!20}{\#\#\# Input:}{\color{gray}\textbackslash n}\begin{CJK}{UTF8}{gkai}
检查情况显示，市场销售的粮油、肉类、水果、蔬菜、蛋奶等生活必需品供应充足，商品价格基本稳定，未发现严重违法违规行为，市场经营秩序总体平稳。\end{CJK}{\color{gray}\textbackslash n\textbackslash n} \\[0.1ex]
& \colorbox{red!20}{\color{jred}\#\#\# Hint:} We prefer to translate it to{\color{gray}\textbackslash n\textbackslash n} \\[0.1ex]
& \colorbox{green!20}{\color{jgreen}\#\#\# Response:}<p>The inspection results showed that there was an adequate supply of daily necessities, including grain, oil, meat, fruit, vegetable, milk, and eggs in the market and commodity prices basically remain stable, the administration found no serious offensive and noncompliant conducts, and the market order remains stable on the whole.</p> rather than <p>The results of the inspection indicate the sufficient supply of living necessities on marketing including cereals and oils, meat, fruits, vegetables, eggs and milk, and the basically stabilized commodity price. The inspection hasn’t found serious violation of laws and regulations. The market order is stable on an overall basis.</p> \\
\midrule
Error-Guided & \colorbox{gray!20}{\#\#\# Instruction:}{\color{gray}\textbackslash n}Translate the following sentences from Chinese to English.{\color{gray}\textbackslash n\textbackslash n} \\[0.3ex]
& \colorbox{gray!20}{\#\#\# Input:}{\color{gray}\textbackslash n}\begin{CJK}{UTF8}{gkai}
检查情况显示，市场销售的粮油、肉类、水果、蔬菜、蛋奶等生活必需品供应充足，商品价格基本稳定，未发现严重违法违规行为，市场经营秩序总体平稳。\end{CJK}{\color{gray}\textbackslash n\textbackslash n} \\[0.3ex]
& \colorbox{red!20}{\color{jred}\#\#\# Hint:} A translation with major accuracy/mistranslation errors could be{\color{gray}\textbackslash n\textbackslash n} \\[0.3ex]
& \colorbox{green!20}{\color{jgreen}\#\#\# Response:}The results of the inspection indicate the sufficient supply of living necessities <v>on marketing</v> including cereals and oils, meat, fruits, vegetables, eggs and milk, and the basically stabilized commodity price. The inspection hasn’t found serious violation of laws and regulations. The market order is stable on an overall basis. \\[0.2ex]
\cdashline{2-2}
& \colorbox{red!20}{\color{jred}\#\#\# Hint:} A translation with minor fluency/grammar errors could be{\color{gray}\textbackslash n\textbackslash n} \\[0.3ex]
& \colorbox{green!20}{\color{jgreen}\#\#\# Response:}The results of the <v>inspection</v> indicate the sufficient supply of living necessities on marketing including cereals and oils, meat, fruits, vegetables, eggs and milk, and the basically stabilized commodity price. The inspection hasn’t found serious violation of laws and regulations. The market order is stable on an overall basis. \\
\bottomrule
\end{tabular}
}
\label{tab:instruction-pool}
\end{table*}

\section{Instruction Pool}
In this section, we introduce the three distinct instruction types: translation instruction, contrastive instruction, and error-guided instruction.
The first instruction guarantees the basic translation ability of LLMs while the latter two regulate the LLMs to align with human-written translation and feedback.

\subsection{Translation Instruction}
\label{sec:translation-instructions}

As traditional translation systems, we rely on bilingual sentence pairs to accomplish the basic translation ability of LLMs.
We follow Stanford Alpaca~\cite{alpaca} to transform bilingual sentence pairs into the instruction-following format, named \textbf{translation instruction}, for finetuning.

Table~\ref{tab:instruction-pool} presents an example of the translation instruction, which includes a preface fixed for all tasks, an ``\#\#\# Instruction:'' to describe the translation task (e.g., stating the language pair), an ``\#\#\# Input:'' with the source sentence, and a ``\#\#\# Response:'' with the target sentence to be generated. 
To ensure the high quality of sentence pairs, we use human-written translations rather than public training data that could be noisy.

\subsection{Contrastive Instruction}
\label{sec:contrastive-instructions}

Besides the basic translation ability, we also want LLMs to understand the relative quality difference between translations. In this way, we may improve the quality of translations by asking LLMs to output the preferred ones.
To realize this goal, we need multiple different translations for each source sentence, which can be acquired by the systems submitted to WMT competitions. Meanwhile, the human evaluation results of these systems also provide scores to reflect the quality differences.

As shown in Table~\ref{tab:instruction-pool}, we form the response by concatenating two translations (e.g., linked by ``rather than''), in which the first translation has a higher quality score. Meanwhile, we indicate that the first translation is preferred in the ``\#\#\# Hint:'' field. 
Essentially, the second translation acts like a negative sample to this sentence pair, which explains the name \textbf{contrastive instruction}.

\subsection{Error-Guided Instruction}
\label{sec:error-guided-instructions}

The potential problem of contrastive instruction is that, it only tells the LLMs that the two translations have quality differences but not clarify which kind of translation errors lead to such differences. However, we want LLMs to learn the correspondence between the errors and the translations. With such a deeper understanding on the translation errors, we may ask LLMs to produce translations with no error so as to improve the quality. 

We propose \textbf{error-guided instruction}. As shown in Table~\ref{tab:instruction-pool}, we use the translation with errors annotated by the ``<v></v>'' span to form the response. Similar to contrastive instruction, we adopt the ``\#\#\# Hint:'' field to indicate the error types. This kind of fine-grained error annotation also comes from the human evaluation data.

\section{Experimental Setups}

\subsection{Training Data}
\label{sec:training-data}

\paragraph{Alpaca Data.} 
This dataset is built by Stanford~Alpaca~\cite{alpaca}\footnote{\url{https://github.com/tatsu-lab/stanford_alpaca}} project, which contains 52.0K instruction-following data of multi-tasks for tuning the LLaMA~\cite{touvron2023llama}\footnote{\url{https://github.com/facebookresearch/llama}} models. 
We call these data \textbf{general~instructions}, which help the resulting ParroT models to maintain capabilities on general tasks.

\paragraph{WMT Validation Data.}
We use human-written validation data from previous WMT competitions rather than public training data to avoid introducing noises into instruction tuning.
In this version, we use the newstest2017-2020 of Chinese$\Leftrightarrow$English~(i.e., Zh$\Leftrightarrow$En) and German$\Leftrightarrow$English~(i.e., De$\Leftrightarrow$En) tasks, which consist of 51.2K sentence pairs for all the four directions. These sentence pairs are formed into the \textbf{translation~instructions}.
    
\paragraph{MQM Human Evaluation Data.} 
Our human feedback data comes from the Multidimensional Quality Metrics (MQM) datasets~\cite{freitag2021experts}\footnote{\url{https://github.com/google/wmt-mqm-human-evaluation}}, which annotate the different translation errors~(e.g., major accuracy/mistranslation, minor fluency/grammar) of top WMT systems.
Due to its higher reliability than Direct Assessment, MQM was introduced to WMT competitions starting from WMT20 but only provided for a few language pairs.  
In this version, we use the MQM data for the WMT20 En$\Rightarrow$De and Zh$\Rightarrow$En submissions. These data are formed into the \textbf{contrastive~instructions} (i.e., 20K) based on the quality scores and the \textbf{error-guided~instructions} (i.e., 26K) based on the error annotations, respectively. 

\paragraph{Automatically Assessed Data.}
Although the Direct Assessment~(DA) data of WMT systems provide scores for language directions that lack MQM data~(i.e., De$\Rightarrow$En, En$\Rightarrow$Zh), we find the DA score to be very unreliable as they could be quite different for two similar translations. Instead, we opt for automatic evaluation metrics like COMET to score the translations of WMT systems. 
We also heuristically determine a rough error level for each translation based on the COMET score, namely, Major Error: [0, 85]; Minor Error: (85, 90]; No Error: (90, 100]. This decision comes in part from the observation that top commercial systems achieve COMET scores of nearly 90 on the Flores subsets~(Table~\ref{tab:mt-performance-llama}). Finally, we obtain 24K contrastive instructions and 29K error-guided instructions.

\textbf{Note}: To obtain a set of diverse instructions, we use the three instructions in~\newcite{jiao2023ischatgpt}, including the one in Table~\ref{tab:instruction-pool}, as the seeds to ask GPT-4~\cite{openai2023gpt4} to paraphrase them. In total, we have 33 different instructions that are randomly combined with the training examples.

\subsection{Model Training}

We conduct our experiments with HuggingFace Transformers\footnote{\url{https://github.com/huggingface/transformers}} on open-source LLMs from both the LLaMA family~\cite{touvron2023llama} and the BLOOM family~\cite{scao2022bloom}. Specifically, we choose LLaMA-7b and BLOOMZ-7b1-mt with matched parameters, and also include LLaMA-13b and BLOOMZ-560m to study the effect of model sizes.
We finetune them to the following variants: 
\begin{itemize}[leftmargin=15pt]
\item \textbf{Alpaca}, as a reimplementation of the Stanford~Alpaca model fine-tuned only on the Alpaca multi-task dataset.
\item \textbf{ParroT-T}, finetuned on the Alpaca multi-task dataset and only the translation instructions from WMT validation data. 
\item \textbf{ParroT}, finetuned on the Alpaca multi-task dataset, and all the three types of instructions introduced above.
\item \textbf{ParroT-LoRA}, finetuned by low-rank adaptation~(LoRA) with default hyper-parameters from \texttt{alpaca-lora}\footnote{\url{https://github.com/tloen/alpaca-lora}}, which results in only 4.2M tunable parameters based on LLaMA-7b.
\end{itemize}
The hyper-parameters for finetuning are basically consistent with Stanford~Alpaca~\cite{alpaca}. 
We finetune the Alpaca and ParroT-T models for 3 epochs on the corresponding data combination. For ParroT and ParroT-LoRA, we finetune them for 1.5 epochs to maintain similar training steps as ParroT-T.
We conduct finetuning on 8 Nvidia A100 GPUs and utilize DeepSpeed\footnote{\url{https://github.com/microsoft/DeepSpeed}}~ZeRO~stage~3 for model parallel.

\subsection{Evaluation}
\label{sec:evaluation}

\paragraph{Test Data.}
We evaluate the translation performance of LLMs on two sources of test sets:
\begin{itemize}[leftmargin=15pt]
    \item \textbf{Flores Subset}: This dataset is a subset of Flores benchmark, in which 50 sentences are sampled for German, English, Romanian and Chinese, respectively, for evaluating the translation performance of ChatGPT~\cite{jiao2023ischatgpt}
    \item \textbf{WMT22 Test Sets}: We also use the test sets from WMT22 competition~\cite{kocmi2022findings}, which are constructed based on more recent content from various domains, including news, social, e-commerce, and conversational domains. The numbers of samples for De$\Rightarrow$En, En$\Rightarrow$De, Zh$\Rightarrow$En and En$\Rightarrow$Zh tasks are 1984, 2037, 1875 and 2037, respectively.
\end{itemize}
For models based on BLOOM, we only evaluate them on WMT22 test sets since the Flores benchmark has been used in the development of BLOOMZ models.   

\paragraph{Metrics.}
For automatic evaluation, we adopt BLEU~\cite{papineni2002bleu} implementated in SacreBLEU~\cite{post2018sacrebleu}\footnote{\url{https://github.com/mjpost/sacrebleu}}, and COMET~\cite{rei2020comet}\footnote{\url{https://github.com/Unbabel/COMET}} from \texttt{Unbabel/wmt22-comet-da}, which are driven by $n$-gram similarity and cross-lingual pretrained models, respectively. 

\begin{table}[t!]
\fontsize{10}{11}\selectfont
\setlength{\tabcolsep}{4pt}
\centering
\caption{Ablation study of key factors on Flores En$\Rightarrow$De subset with Alpaca based on LLaMA-7b.}
\begin{threeparttable}
\begin{tabular}{ccc rc}
\toprule
\bf Prompt & \bf Instruct. & \bf Search & \bf BLEU & \bf COMET \\
\midrule
\multirow{4}{*}{\texttt{no-input}} & 
\multirow{2}{*}{\textsc{Tp1}} & sample & 20.0 & 80.0 \\
 &  & beam~4 & 22.1 & 79.1 \\
 & \multirow{2}{*}{\textsc{Tp3}} & sample & 19.4 & 79.0 \\
 &  & beam~4 & 21.5 & 79.0 \\
\hline
\multirow{4}{*}{\texttt{input}} & \multirow{2}{*}{\textsc{Tp1}} & sample & 21.0 & 79.5 \\
 &  & beam~4 & \bf 23.3 & \bf 80.5 \\
 \cdashline{2-5}
 & \multirow{2}{*}{\textsc{Tp3}} & sample & 19.3 & 78.6 \\
 &  & beam~4 & 20.6 & 80.0 \\
\bottomrule
\end{tabular}
\end{threeparttable}
\label{tab:ablation-study}
\end{table}

\begin{table*}[t!]
\fontsize{10}{11}\selectfont
\centering
\caption{Translation performance of LLaMA models on Flores subsets and WMT22 test sets.}
\begin{threeparttable}
\begin{tabular}{l rcrcrcrc}
\toprule
\multirow{2}{*}{\bf System} & \multicolumn{2}{c}{\bf De$\Rightarrow$En} & \multicolumn{2}{c}{\bf En$\Rightarrow$De} & \multicolumn{2}{c}{\bf Zh$\Rightarrow$En} & \multicolumn{2}{c}{\bf En$\Rightarrow$Zh} \\
\cmidrule(lr){2-3}\cmidrule(lr){4-5}\cmidrule(lr){6-7}\cmidrule(lr){8-9}
& BLEU & COMET & BLEU & COMET & BLEU & COMET & BLEU & COMET \\
\midrule
\multicolumn{9}{c}{\bf Flores Subsets}\\
Google & 45.0 & 88.7 & 41.1 & 88.6 & \bf 31.6 & \bf 87.7 & 43.5 & \bf 88.4 \\
DeepL & \bf 49.2 & \bf 89.7 & 41.4 & 89.0 & 31.2 & 87.3 & \bf 44.3 & 88.1 \\
ChatGPT & 43.7 & 89.1 & 38.8 & 88.1 & 24.7 & 85.8 & 38.2 & 86.9 \\
GPT-4 & 46.0 & 89.3 & \bf 45.7 & \bf 89.2 & 28.5 & 87.4 & 42.5 & 88.4 \\
\hline
\multicolumn{9}{c}{\it Base Model: LLaMA-7b} \\
Vanilla & 3.4 & 60.1 & 2.4 & 49.0 & 1.8 & 53.7 & 0.1 & 47.6 \\
Alpaca & 36.6 & 86.8 & 23.3 & 80.5 & 15.1 & 81.2 & 9.8 & 58.6 \\
Alpaca-LoRA & 40.7 & 87.7 & 24.6 & 84.0 & 16.4 & 81.5 & 14.5 & 70.5 \\
\hdashline
ParroT-T & 41.3 & 87.7 & 28.5 & 83.3 & 19.5 & 83.1 & 24.7 & 79.9 \\
ParroT & 41.0 & 87.9 & 30.8 & 84.3 & 19.2 & \bf 83.9 & 25.8 & 80.1 \\
~~~+ Infer w/ Prefer. & 38.1 & 87.6 & 23.0 & 83.9 & 18.6 & 83.1 & 22.5 & 80.1 \\
~~~+ Infer w/ No Err. & 42.2 & \bf 88.7 & \bf 32.1 & 84.9 & \bf 21.5 & 83.7 & \bf27.4 & \bf 81.8 \\
\hdashline
ParroT-LoRA & \bf 43.8 & 88.3 & 29.0 & 84.9 & 16.9 & 80.6 & 14.8 & 71.5 \\
~~~+ Infer w/ No Err. & 42.0 & 88.0 & 29.8 & \bf 85.4 & 17.4 & 81.3 & 19.8 & 76.7 \\
\midrule
\midrule
\multicolumn{9}{c}{\bf WMT22 Test Sets} \\
Google & 33.3 & 84.8 & \bf 38.4 & 87.1 & \bf 28.6 & 80.9 & \bf 49.9 & \bf 87.4 \\
DeepL & 32.8 & 84.7 & 36.2 & \bf 87.9 & 24.2 & 79.3 & 44.5 & 86.4 \\
GPT-4 & \bf 33.4 & \bf 84.9 & 34.5 & 87.4 & 24.8 & \bf 82.3 & 41.3 & 87.0 \\
\hline
\multicolumn{9}{c}{\it Base Model: LLaMA-7b} \\
Vanilla & 2.9 & 52.8 & 1.6 & 45.3 & 1.2 & 50.3 & 0.3 & 46.3 \\
Alpaca & 27.8 & 82.3 & 20.1 & 78.1 & 14.2 & 74.0 & 10.4 & 62.1 \\
Alpaca-LoRA & 28.9 & \bf 83.2 & 22.1 & 81.3 & 16.1 & 75.6 & 16.3 & 70.6 \\
\hdashline
ParroT-T & 26.6 & 82.5 & 24.0 & 80.4 & 18.1 & 75.3 & 27.0 & 78.4 \\
ParroT & 27.3 & 82.4 & 24.6 & 81.2 & 18.9 & 75.2 & 28.1 & 79.3 \\
~~~+ Infer w/ No Err. & 27.3 & 82.4 & \bf 26.1 & 81.6 & \bf 20.2 & \bf 75.9 & \bf 30.3 & \bf 80.3 \\
\hdashline
ParroT-LoRA & 28.8 & 82.8 &  24.0 & 81.4 & 18.2 & 74.7 & 19.9 & 73.7 \\
~~~+ Infer w/ No Err. & \bf 29.8 & 83.0 & 24.8 & \bf 81.6 & 19.2 & 75.0 & 20.7 & 74.5 \\
\hline
\multicolumn{9}{c}{\it Base Model: LLaMA-13b} \\
Alpaca & 29.7 & 83.1 & 21.4 & 79.4 & 16.2 & 75.9 & 17.6 & 70.8 \\
\hdashline
ParroT & 27.6 & 83.2 & 27.0 & \bf 82.8 & 19.9 & 75.8 & 30.9 & \bf 81.1 \\
~~~+ Infer w/ No Err. & \bf 31.1 & \bf 83.6 & \bf 28.1 & 82.6 & \bf 21.7 & \bf 76.7 & \bf 31.7 & 81.0 \\
\bottomrule
\end{tabular}
\end{threeparttable}
\label{tab:mt-performance-llama}
\end{table*}

\section{Results}

\subsection{Ablation Study}

Before diving into more experiments, we investigate  some factors that may affect the translation performance of LLMs. By default, we conduct the ablation studies on the Flores En$\Rightarrow$De subset with the Alpaca model based on LLaMA-7b.

\paragraph{Prompt Format.}
In the Alpaca multi-task dataset, about 60\% examples contain empty ``\#\#\# Input:'', which results in two different prompt formats during finetuning, i.e., \texttt{prompt-input} and \texttt{prompt-no-input}. During inference, they use \texttt{prompt-no-input} which combines the instruction and input to fill the ``\#\#\# Instruction:'' field, introducing the inconsistency between finetuning and inference. Therefore, we study if such an operation makes any performance variation.

\paragraph{Instruction Variation.}
Recent studies~\cite{jiao2023ischatgpt,zhang2023prompting} suggest that LLMs are sensitive to task instructions, which could vary the translation performance considerably. We conduct a brief study for this by comparing the \textsc{Tp1} and \textsc{Tp3} instructions in \newcite{jiao2023ischatgpt}. \textsc{Tp1} is the one presented in Table~\ref{tab:instruction-pool} while \textsc{Tp3} is ``\texttt{Please provide the [TGT] translation for the following sentences.}'', which was demonstrated a better choice when tested on ChatGPT\footnote{\url{https://chat.openai.com}}.

\paragraph{Search Algorithm.}
In machine translation, the beam search strategy~\cite{sutskever2014seq2seq,freitag2017beam,vaswani2017attention} has been the standard search algorithm for inference. 
However, beam search requires high computation costs which becomes infeasible with the LLMs, since they can easily induce out-of-memory~(OOM) issues. Therefore, more efficient search algorithms such as sampling may have to be the choice.
Therefore, we compare the sampling strategy~\cite{alpaca} and the beam search strategy with a beam size of 4 for this factor.

\vspace{5pt}
Table~\ref{tab:ablation-study} presents the results of these ablation studies. We have the following observations: (1) The \texttt{prompt-input} performs slightly better than \texttt{prompt-no-input} though the gap is marginal. (2) The \textsc{Tp1} instruction works better on Alpaca than \textsc{Tp3} which is different from that on ChatGPT. (3) Generally, beam search outperforms sampling significantly, especially in terms of BLEU score. 
Therefore, we use \texttt{prompt-input} + \textsc{Tp1} + beam search as the default setting for inference.

\begin{table*}[ht]
\fontsize{10}{11}\selectfont
\centering
\caption{Translation performance of BLOOM models on WMT22 test sets.}
\begin{threeparttable}
\begin{tabular}{l rcrcrcrc}
\toprule
\multirow{2}{*}{\bf System} & \multicolumn{2}{c}{\bf De$\Rightarrow$En} & \multicolumn{2}{c}{\bf En$\Rightarrow$De} & \multicolumn{2}{c}{\bf Zh$\Rightarrow$En} & \multicolumn{2}{c}{\bf En$\Rightarrow$Zh} \\
\cmidrule(lr){2-3}\cmidrule(lr){4-5}\cmidrule(lr){6-7}\cmidrule(lr){8-9}
& BLEU & COMET & BLEU & COMET & BLEU & COMET & BLEU & COMET \\
\midrule
\multicolumn{9}{c}{\it Base Model: BLOOMZ-560m} \\
Alpaca & 4.4 & 55.2 & 0.5 & 30.8 & 6.9 & 70.1 & 2.0 & 54.0 \\
\hdashline
ParroT & 16.4 & 68.9 & 13.3 & 57.7 & 16.0 & 74.8 & 25.4 & 79.0 \\
~~~+ Infer w/ No Err. & \bf 16.9 & \bf 69.3 & 12.8 & 56.8 & 15.7 & \bf 75.0 & \bf 26.3 & \bf 79.5 \\
\hline
\multicolumn{9}{c}{\it Base Model: BLOOMZ-7b1-mt} \\
Alpaca & 17.6 & 73.0 & 3.1 & 44.5 & 13.0 & 76.4 & 23.9 & 81.8 \\
\hdashline
ParroT & 23.1 & 77.6 & 20.0 & 72.7 & 21.4 & 78.5 & 32.4 & \bf 83.6 \\
~~~+ Infer w/ No Err. & \bf 24.9 & \bf 78.0 & \bf 20.5 & \bf 73.6 & \bf 22.7 & \bf 79.0 & \bf 34.5 & 83.5 \\
\bottomrule
\end{tabular}
\end{threeparttable}
\label{tab:mt-performance-bloom}
\end{table*}

\subsection{Main Results}

Table~\ref{tab:mt-performance-llama} and Table~\ref{tab:mt-performance-bloom} present the translation performance of LLaMA and BLOOM models on the test sets. For Flores subsets, we include the baseline results reported in \newcite{jiao2023ischatgpt}.

\paragraph{Instruction tuning exploits the potential of vanilla LLMs for machine translation.}
Table~\ref{tab:mt-performance-llama} shows that the vanilla LLaMA-7b without any further training performs badly on the Flores subsets. By inspecting the outputs, we find that the vanilla LLaMA-7b model tends to generate very long sentences (e.g., copy the instructions, continuing text expansion), which makes the generated text not faithful to the source sentences and also not grammatically correct. The reason could be the long context modeling during pretraining. Another reason is that we use the Alpaca inference format, which is basically a zero-shot setting that exhibits no guidance for translation. 

Tuning LLaMA-7b on the Alpaca multi-task dataset~(i.e., Alpaca) can ameliorate the above issue, resulting in complete generations with proper lengths. 
We find that Alpaca performs much better on translation, which may benefit from the 0.5\% translation instructions in the Alpaca multi-task dataset.
However, the best performance is mainly observed on high-resource directions like De$\Rightarrow$En, due to the dominant language of Alpaca dataset in English.
Further introducing a small amount of translation instructions (i.e., ParroT-T) in the four language directions can significantly improve the performance, especially for En$\Rightarrow$Zh, in which Chinese was unseen in the pretraining of LLaMA models~\cite{touvron2023llama}.
The findings of these LLaMA-based models are also consistent with that on the WMT22 test sets.

\paragraph{Learning from low-quality translations annotated by humans is also important.}
While presenting the high-quality bilingual pairs to LLMs is important, as discussed above, we argue that low-quality translations annotated by humans also bring benefits.
As shown in Table~\ref{tab:mt-performance-llama}, without hint in inference, ParroT outperforms ParroT-T slightly on translation directions from English to other languages (i.e., En$\Rightarrow$De, En$\Rightarrow$Zh). However, when asking ParroT to generate translations \textbf{with no error}, the performance can be significantly improved across translation directions and test sets. We speculate that ParroT does learn the relationship between errors and translations by error-guided instruction, such that it can avoid the translation errors as much as possible when the hint of no error is provided.

A bit unexpected is that when asking ParroT to generate preferred translations, the performance drops considerably. 
As stated in Section~\ref{sec:error-guided-instructions}, contrastive instruction only indicates that two translations may have quality differences but not state why, which is difficult for LLMs to identify by themselves. Previous study by~\newcite{min2022rethinking} also suggests that it is easier for LLMs to learn the instruction formats rather than the input-response patterns, which may explain the phenomenon here.

\begin{table}[t]
\fontsize{10}{11}\selectfont
\setlength{\tabcolsep}{5pt}
\centering
\caption{Effects of error levels as hints during inference. \colorbox{jred!50}{Red}: improvement; \colorbox{jgreen!50}{Green}: degradation. }
\begin{threeparttable}
\begin{tabular}{l cc cc cc cc}
\toprule
\multirow{2}{*}{\bf Hint} & \multicolumn{2}{c}{\bf En$\Rightarrow$De} & \multicolumn{2}{c}{\bf Zh$\Rightarrow$En} \\
\cmidrule(lr){2-3}\cmidrule(lr){4-5}
& BLEU & COMET & BLEU & COMET \\
\midrule
None & 30.8 & 84.3 & 19.2 & 83.9 \\
\hline
No Err. & \colorbox{jred!26}{32.1} & \colorbox{jred!8}{84.9} & \colorbox{jred!46}{21.5} & \colorbox{jgreen!4}{83.7} \\
Minor Err. & \colorbox{jgreen!40}{28.8} & \colorbox{jgreen!14}{83.6} & \colorbox{jred!28}{20.6} & \colorbox{jgreen!36}{82.1} \\
Major Err. & \colorbox{jgreen!46}{28.5} & \colorbox{jgreen!28}{82.9} & \colorbox{jred!2}{19.3} & \colorbox{jgreen!68}{80.5} \\
\bottomrule
\end{tabular}
\end{threeparttable}
\label{tab:analysis-EGI}
\end{table}

\begin{table*}[t]
\fontsize{10}{11}\selectfont
\centering
\caption{Examples from Flores Zh$\Rightarrow$En subset when translated with varied error levels.
\colorbox{red!50}{Red}: translation errors.}
\resizebox{0.99\textwidth}{!}{
\begin{tabular}{l c p{14cm}}
\toprule
& \bf COMET & \multicolumn{1}{c}{\bf Example} \\
\midrule
\textsc{Src} & n/a & \begin{CJK}{UTF8}{gkai}
例如，在卢瓦尔河谷和莱茵河谷参观城堡，或者乘游轮去游览多瑙河沿岸好玩的城市，或者沿伊利运河乘船游玩。\end{CJK}\\
\textsc{Ref} & n/a & For example visiting castles in the Loire Valley, the Rhine valley or taking a cruise to interesting cites on the Danube or boating along the Erie Canal. \\
\hline
None & 81.2 & For example, visiting the castles in the \colorbox{red!20}{Rhine Valley} and the \colorbox{red!20}{Lahn Valley}, or taking a cruise along the \colorbox{red!20}{Rhine River} to explore the cities \fcolorbox{red!100}{red!50}{along the river}, or taking a boat trip along the \colorbox{red!20}{Illi River}. \\
No Err. & 81.4 & For example, visiting the castles in the \colorbox{red!20}{Rhine Valley} and the \colorbox{red!20}{Ruhr Valley}, or taking a cruise to visit the cities along the \colorbox{red!20}{Rhine River}, or taking a boat ride along the \colorbox{red!20}{Illi River}. \\
Minor Err. & 78.1 & For example, visiting the castles in the \colorbox{red!20}{Rhine Valley} and the \colorbox{red!20}{Ruhr Valley}, or taking a cruise to visit the cities along the \colorbox{red!20}{Rhine River}, or taking a boat to \fcolorbox{red!100}{red!50}{play in the cities} along the \colorbox{red!20}{Elbe River}. \\
Major Err. & 65.3 & For example, visiting the \fcolorbox{red!100}{red!50}{fortress} in the valley of the \colorbox{red!20}{River Volga} and the \colorbox{red!20}{River Levan}, or taking a cruise to visit the cities along the \colorbox{red!20}{River Volga}, or taking a boat to \fcolorbox{red!100}{red!50}{play in the cities} along the \colorbox{red!20}{River Volga}. \\
\bottomrule
\end{tabular}
}
\label{tab:analysis-EGI-case}
\end{table*}

\paragraph{Parameter efficient finetuning may prevent LLMs from overfitting.}
We also try low-rank adaptation~\cite[LoRA,][]{hu2022lora} to finetune partial parameters of LLMs for efficiency.
Experimental results in Table~\ref{tab:mt-performance-llama} show that Alpaca-LoRA outperforms its full model counterpart noticeably. We speculate that LoRA can prevent LLMs from overfitting the small Alpaca multi-task dataset, leading to a stronger generalization ability. 
However, applying LoRA to ParroT exhibits distinct behaviors for high-resource and low-resource translation directions. Specifically, ParroT-LoRA outperforms the corresponding full model ParroT on De$\Rightarrow$En but performs much worse on the other directions. 
It seems that the small amount of tunable parameters also hinder the learning of instructions from other translation directions. Obviously, the hyper-parameters of LoRA should also be properly adjusted to better learn from more instruction data.

\paragraph{LLMs families and sizes also matter.}
For both LLaMA and BLOOM families, larger models can achieve much better translation performance after instruction tuning. Our ParroT framework proves to be effective across all the models.
Comparing the two LLMs families, the ParroT model based on BLOOMZ-7b1-mt performs much better on Zh$\Rightarrow$En and En$\Rightarrow$Zh directions than those based on LLaMA-7b, which mainly results from the better modeling of Chinese during the pretraining process of BLOOM.

\begin{table}[t!]
\fontsize{10}{11}\selectfont
\setlength{\tabcolsep}{5pt}
\centering
\caption{Effects of preference as hints during inference.
\colorbox{jred!50}{Red}: improvement; \colorbox{jgreen!50}{Green}: degradation.}
\begin{threeparttable}
\begin{tabular}{l cc cc cc cc}
\toprule
\multirow{2}{*}{\bf Hint} & \multicolumn{2}{c}{\bf En$\Rightarrow$De} & \multicolumn{2}{c}{\bf Zh$\Rightarrow$En} \\
\cmidrule(lr){2-3}\cmidrule(lr){4-5}
& BLEU & COMET & BLEU & COMET \\
\midrule
None & 30.8 & 84.3 & 19.2 & 83.9 \\
\hline
Prefer. & \colorbox{jgreen!100}{23.0} & \colorbox{jgreen!8}{83.9} & \colorbox{jgreen!12}{18.6} & \colorbox{jgreen!12}{83.1}  \\
Unprefer. & \colorbox{jgreen!34}{29.1} & \colorbox{jgreen!12}{83.7} & \colorbox{jred!8}{19.6} & \colorbox{jgreen!32}{82.3} \\
\bottomrule
\end{tabular}
\end{threeparttable}
\label{tab:analysis-CI}
\end{table}

\paragraph{Automatic evaluation tools can be effective in constructing error-guided instructions.}
In Section~\ref{sec:training-data}, we construct the automatically assessed data for De$\Rightarrow$En and En$\Rightarrow$Zh that are not provided with the MQM data. 
As shown in Table~\ref{tab:mt-performance-llama} and Table~\ref{tab:mt-performance-bloom}, we can observe considerable improvements of error-guided instruction on these two translation directions.
It demonstrates the potential of automatic evaluation tools (i.e., COMET) in providing the quality information of translations, as an augmentation to translation directions that lack human annotation data. 

\subsection{Analysis}

We conduct more analyses to understand the effects of our instruction types. By default, we use the model variants based on LLaMA-7b, and the Flores subsets.

\paragraph{Effectiveness of Error-Guided Instruction.}
To understand how error-guided instruction works, we investigate the behavior of ParroT when asking it to generate translations with varied error levels as hints.
As shown in Table~\ref{tab:analysis-EGI}, the translation quality is getting worse from no error to minor error to major error, especially in terms of COMET score.
The translations generated with no hint are usually comparable with the minor error level. 
It demonstrates that ParroT can place erroneous translations into other locations of the probability space with the regulation of human annotations.
As a result, ParroT is more likely to generate high-quality translation with ``no error''.

For qualitative analysis, we show an example from Flores Zh$\Rightarrow$En subset in Table~\ref{tab:analysis-EGI-case}, in which we highlight all errors in each translation. Compared to no error level, minor and major error levels tend to produce more over-translations and mis-translations.
It is also important to point out that no error level does not guarantee that completely correct translations will be generated, especially for named entities, which we attribute to the under-explored translation abilities of current LLMs.

\paragraph{Failure of Contrastive Instruction.}
We try to understand why contrastive instruction does not work. By examining the responses of ParroT when asking it to generate preferred translations, we observe significant differences in lexical choices between the ``preferred'' and ``unpreferred''~(i.e., the second translation in the response) translations. 
Surprisingly, as shown in Table~\ref{tab:analysis-CI}, the ``unpreferred'' translations obtain a much higher BLEU score but the situation is different for the COMET score. It indicates that ParroT attempted to identify the quality differences between the first and second translations in the contrastive instructions through lexical choices, which is a low-level pattern to reflect the translation quality. 
One potential reason is that the WMT systems are so competitive with each other that the quality differences between them are too subtle for the LLM to learn effectively.
We will investigate more about contrastive instruction in future work.

\section{Related Work}

\paragraph{LLMs for MT.}
With the increasing capacity of LLMs, they have become good few-shot learners~\cite{brown2020gpt3,lin2022few} on various NLP tasks, including machine translation.
A number of recent studies focus on how to prompt LLMs for machine translation, including prompt template comparison~\cite{zhang2023prompting}, few-shot example selection~\cite{agrawal2022context,vilar2022prompting}, domain adaptation~\cite{moslem2023adaptive}, and rare word translation~\cite{ghazvininejad2023dictionary}.
However, our ParroT framework aims to develop instant translation capability for chatbots without few-shot examples.
This is consistent with the performance of ChatGPT and GPT-4~\cite{openai2023gpt4}, which exhibit excellent translation ability~\cite{jiao2023ischatgpt,bang2023M3ChatGPT,he2023maps,liang2023mad} during chat.

\paragraph{Instruction Tuning.}
To eliminate the reliance on few-shot examples, recent studies also try to finetune LLMs on a small amount of instructions covering different NLP tasks, making the LLMs zero-shot learners~\cite{mishra2022cross,wei2022finetuned}. With the emergence of various powerful open-source LLMs such as BLOOM~\cite{scao2022bloom} and LLaMA~\cite{touvron2023llama}, there has been a boom for creating instruction data and tuning customized chatbots, for example, Alpaca~\cite{alpaca}, Vicuna, WizardLM~\cite{xu2023wizardlm} and the like. 
However, most of these studies focus on developing chatbots that are capable of general NLP tasks, while we pay more attention to machine translation.
More importantly, apart from the instructions built from parallel translation data, we also transform human feedback data into instructions and demonstrate its effectiveness in improving the translation performance.

\section{Conclusion}
We propose ParroT to enhance and regulate the translation abilities during chat based on open-source LLMs, human-written translation and feedback data.
We reformulate translation data into the instruction-following style, and introduce a “Hint” field for incorporating extra requirements to regulate the translation process. Accordingly, we propose three instruction types for finetuning ParroT models, i.e., translation instruction, contrastive instruction, and error-guided instruction.
Experiments on Flores subsets and WMT22 test sets suggest that translation instruction improves the translation performance of vanilla LLMs significantly while error-guided instruction can lead to further improvement, demonstrating the importance of learning from low-quality translations annotated by humans.
While we only use three instruction types in this paper, it is natural to extend ParroT to other hints (e.g., entity alignments), which we leave for future exploration.

\section*{Limitations}

This work performs a preliminary exploration on the instant translation capability for chatbots, which can be further improved in the following aspects:
\begin{itemize}[leftmargin=10pt]
    \item \textbf{Instruction Variants}:
    Presently, the instructions only support the translation of incoming sentences. It may be beneficial for chatbots to also translate previous chat records when users struggle to comprehend responses in foreign languages.
    \item \textbf{Contrastive Translations}:
    In this study, we did not observe performance improvements related to contrastive instructions, possibly due to incorrect instruction formatting. By exploring alternative formats, such as automatic post-editing (APE), we could potentially capitalize on the advantages of contrastive translations.
    \item \textbf{LoRA Effectiveness}:
    The current analysis did not reveal consistent performance improvements when using LoRA as compared to full model training. It may be necessary to adjust the number of tunable parameters according to the dataset size for better results.
\end{itemize}



\bibliography{anthology,custom}

\begin{thebibliography}{39}
\expandafter\ifx\csname natexlab\endcsname\relax\def\natexlab#1{#1}\fi

\bibitem[{Agrawal et~al.(2022)Agrawal, Zhou, Lewis, Zettlemoyer, and
  Ghazvininejad}]{agrawal2022context}
Sweta Agrawal, Chunting Zhou, Mike Lewis, Luke Zettlemoyer, and Marjan
  Ghazvininejad. 2022.
\newblock In-context examples selection for machine translation.
\newblock \emph{arXiv}.

\bibitem[{Bang et~al.(2023)Bang, Cahyawijaya, Lee, Dai, Su, Wilie, Lovenia, Ji,
  Yu, Chung, Do, Xu, and Fung}]{bang2023M3ChatGPT}
Yejin Bang, Samuel Cahyawijaya, Nayeon Lee, Wenliang Dai, Dan Su, Bryan Wilie,
  Holy Lovenia, Ziwei Ji, Tiezheng Yu, Willy Chung, Quyet~V. Do, Yan Xu, and
  Pascale Fung. 2023.
\newblock A multitask, multilingual, multimodal evaluation of {ChatGPT} on
  reasoning, hallucination, and interactivity.

\bibitem[{Brown et~al.(2020)Brown, Mann, Ryder, Subbiah, Kaplan, Dhariwal,
  Neelakantan, Shyam, Sastry, Askell et~al.}]{brown2020gpt3}
Tom Brown, Benjamin Mann, Nick Ryder, Melanie Subbiah, Jared~D Kaplan, Prafulla
  Dhariwal, Arvind Neelakantan, Pranav Shyam, Girish Sastry, Amanda Askell,
  et~al. 2020.
\newblock Language models are few-shot learners.
\newblock \emph{NeurIPS}.

\bibitem[{Farhad et~al.(2021)Farhad, Arkady, Magdalena, Ond{\v{r}}ej, Rajen,
  Vishrav, Costa-jussa, Cristina, Angela, Christian et~al.}]{farhad2021wmt}
Akhbardeh Farhad, Arkhangorodsky Arkady, Biesialska Magdalena, Bojar
  Ond{\v{r}}ej, Chatterjee Rajen, Chaudhary Vishrav, Marta~R Costa-jussa,
  Espa{\~n}a-Bonet Cristina, Fan Angela, Federmann Christian, et~al. 2021.
\newblock Findings of the 2021 conference on machine translation ({WMT21}).
\newblock In \emph{WMT}.

\bibitem[{Freitag and Al-Onaizan(2017)}]{freitag2017beam}
Markus Freitag and Yaser Al-Onaizan. 2017.
\newblock Beam search strategies for neural machine translation.
\newblock \emph{ACL}.

\bibitem[{Freitag et~al.(2021)Freitag, Foster, Grangier, Ratnakar, Tan, and
  Macherey}]{freitag2021experts}
Markus Freitag, George Foster, David Grangier, Viresh Ratnakar, Qijun Tan, and
  Wolfgang Macherey. 2021.
\newblock Experts, errors, and context: A large-scale study of human evaluation
  for machine translation.
\newblock \emph{TACL}.

\bibitem[{Ghazvininejad et~al.(2023)Ghazvininejad, Gonen, and
  Zettlemoyer}]{ghazvininejad2023dictionary}
Marjan Ghazvininejad, Hila Gonen, and Luke Zettlemoyer. 2023.
\newblock Dictionary-based phrase-level prompting of large language models for
  machine translation.
\newblock \emph{arXiv}.

\bibitem[{He et~al.(2023)He, Liang, Jiao, Zhang, Yang, Wang, Tu, Shi, and
  Wang}]{he2023maps}
Zhiwei He, Tian Liang, Wenxiang Jiao, Zhuosheng Zhang, Yujiu Yang, Rui Wang,
  Zhaopeng Tu, Shuming Shi, and Xing Wang. 2023.
\newblock Exploring human-like translation strategy with large language models.
\newblock \emph{arXiv}.

\bibitem[{Hou et~al.(2022)Hou, Jiao, Liu, Allen, Tu, and
  Sachan}]{hou2022adapters}
Yifan Hou, Wenxiang Jiao, Meizhen Liu, Carl Allen, Zhaopeng Tu, and Mrinmaya
  Sachan. 2022.
\newblock Adapters for enhanced modeling of multilingual knowledge and text.
\newblock In \emph{Findings of the Association for Computational Linguistics:
  EMNLP 2022}, pages 3902--3917.

\bibitem[{Hu et~al.(2022)Hu, Wallis, Allen-Zhu, Li, Wang, Wang, Chen
  et~al.}]{hu2022lora}
Edward~J Hu, Phillip Wallis, Zeyuan Allen-Zhu, Yuanzhi Li, Shean Wang, Lu~Wang,
  Weizhu Chen, et~al. 2022.
\newblock {LoRA}: Low-rank adaptation of large language models.
\newblock In \emph{ICLR}.

\bibitem[{Jiao et~al.(2022)Jiao, Tu, Li, Wang, Huang, and
  Shi}]{jiao2022tencent}
Wenxiang Jiao, Zhaopeng Tu, Jiarui Li, Wenxuan Wang, Jen-tse Huang, and Shuming
  Shi. 2022.
\newblock Tencent's multilingual machine translation system for {WMT22}
  large-scale african languages.
\newblock In \emph{WMT}.

\bibitem[{Jiao et~al.(2023)Jiao, Wang, tse Huang, Wang, and
  Tu}]{jiao2023ischatgpt}
Wenxiang Jiao, Wenxuan Wang, Jen tse Huang, Xing Wang, and Zhaopeng Tu. 2023.
\newblock {I}s {ChatGPT} a good translator? {Yes} with {GPT-4} as the engine.
\newblock In \emph{ArXiv}.

\bibitem[{Johnson et~al.(2017)Johnson, Schuster, Le, Krikun, Wu, Chen, Thorat,
  Vi{\'e}gas, Wattenberg, Corrado et~al.}]{johnson2017google}
Melvin Johnson, Mike Schuster, Quoc~V Le, Maxim Krikun, Yonghui Wu, Zhifeng
  Chen, Nikhil Thorat, Fernanda Vi{\'e}gas, Martin Wattenberg, Greg Corrado,
  et~al. 2017.
\newblock Google’s multilingual neural machine translation system: Enabling
  zero-shot translation.
\newblock \emph{TACL}.

\bibitem[{Kocmi et~al.(2022)Kocmi, Bawden, Bojar, Dvorkovich, Federmann,
  Fishel, Gowda, Graham, Grundkiewicz, Haddow et~al.}]{kocmi2022findings}
Tom Kocmi, Rachel Bawden, Ond{\v{r}}ej Bojar, Anton Dvorkovich, Christian
  Federmann, Mark Fishel, Thamme Gowda, Yvette Graham, Roman Grundkiewicz,
  Barry Haddow, et~al. 2022.
\newblock Findings of the 2022 conference on machine translation ({WMT22}).
\newblock In \emph{WMT}.

\bibitem[{Liang et~al.(2023)Liang, He, Jiao, Wang, Wang, Wang, Yang, Tu, and
  Shi}]{liang2023mad}
Tian Liang, Zhiwei He, Wenxiang Jiao, Xing Wang, Yan Wang, Rui Wang, Yujiu
  Yang, Zhaopeng Tu, and Shuming Shi. 2023.
\newblock Encouraging divergent thinking in large language models through
  multi-agent debate.
\newblock \emph{arXiv}.

\bibitem[{Lin et~al.(2022)Lin, Mihaylov, Artetxe, Wang, Chen, Simig, Ott,
  Goyal, Bhosale, Du et~al.}]{lin2022few}
Xi~Victoria Lin, Todor Mihaylov, Mikel Artetxe, Tianlu Wang, Shuohui Chen,
  Daniel Simig, Myle Ott, Naman Goyal, Shruti Bhosale, Jingfei Du, et~al. 2022.
\newblock Few-shot learning with multilingual generative language models.
\newblock In \emph{EMNLP}.

\bibitem[{Liu et~al.(2023)Liu, Sferrazza, and Abbeel}]{liu2023CoH}
Hao Liu, Carmelo Sferrazza, and Pieter Abbeel. 2023.
\newblock Chain of hindsight aligns language models with feedback.
\newblock \emph{arXiv}.

\bibitem[{Min et~al.(2022)Min, Lyu, Holtzman, Artetxe, Lewis, Hajishirzi, and
  Zettlemoyer}]{min2022rethinking}
Sewon Min, Xinxi Lyu, Ari Holtzman, Mikel Artetxe, Mike Lewis, Hannaneh
  Hajishirzi, and Luke Zettlemoyer. 2022.
\newblock Rethinking the role of demonstrations: What makes in-context learning
  work?
\newblock In \emph{EMNLP}.

\bibitem[{Mishra et~al.(2022)Mishra, Khashabi, Baral, and
  Hajishirzi}]{mishra2022cross}
Swaroop Mishra, Daniel Khashabi, Chitta Baral, and Hannaneh Hajishirzi. 2022.
\newblock Cross-task generalization via natural language crowdsourcing
  instructions.
\newblock In \emph{ACL}.

\bibitem[{Moslem et~al.(2023)Moslem, Haque, and Way}]{moslem2023adaptive}
Yasmin Moslem, Rejwanul Haque, and Andy Way. 2023.
\newblock Adaptive machine translation with large language models.
\newblock \emph{arXiv}.

\bibitem[{Omar et~al.(2023)Omar, Mangukiya, Kalnis, and
  Mansour}]{Omar2023ChatGPTVT}
Reham Omar, Omij Mangukiya, Panos Kalnis, and Essam Mansour. 2023.
\newblock {ChatGPT} versus traditional question answering for knowledge graphs:
  Current status and future directions towards knowledge graph chatbots.
\newblock \emph{arXiv}.

\bibitem[{OpenAI(2023)}]{openai2023gpt4}
OpenAI. 2023.
\newblock {GPT-4} technical report.
\newblock \emph{arXiv}.

\bibitem[{Ouyang et~al.(2022)Ouyang, Wu, Jiang, Almeida, Wainwright, Mishkin,
  Zhang, Agarwal, Slama, Ray et~al.}]{ouyang2022InstructGPT}
Long Ouyang, Jeff Wu, Xu~Jiang, Diogo Almeida, Carroll~L Wainwright, Pamela
  Mishkin, Chong Zhang, Sandhini Agarwal, Katarina Slama, Alex Ray, et~al.
  2022.
\newblock Training language models to follow instructions with human feedback.
\newblock \emph{arXiv}.

\bibitem[{Pal et~al.(2016)Pal, Naskar, Vela, and van Genabith}]{pal2016neural}
Santanu Pal, Sudip~Kumar Naskar, Mihaela Vela, and Josef van Genabith. 2016.
\newblock A neural network based approach to automatic post-editing.
\newblock In \emph{ACL}.

\bibitem[{Papineni et~al.(2002)Papineni, Roukos, Ward, and
  Zhu}]{papineni2002bleu}
Kishore Papineni, Salim Roukos, Todd Ward, and Wei-Jing Zhu. 2002.
\newblock {BLEU}: {A} method for automatic evaluation of machine translation.
\newblock In \emph{ACL}.

\bibitem[{Post(2018)}]{post2018sacrebleu}
Matt Post. 2018.
\newblock A call for clarity in reporting {BLEU} scores.
\newblock In \emph{WMT}.

\bibitem[{Rei et~al.(2020)Rei, Stewart, Farinha, and Lavie}]{rei2020comet}
Ricardo Rei, Craig Stewart, Ana~C Farinha, and Alon Lavie. 2020.
\newblock {COMET}: A neural framework for {MT} evaluation.
\newblock In \emph{EMNLP}.

\bibitem[{Scao et~al.(2022)Scao, Fan, Akiki, Pavlick, Ili{\'c}, Hesslow,
  Castagn{\'e}, Luccioni, Yvon, Gall{\'e} et~al.}]{scao2022bloom}
Teven~Le Scao, Angela Fan, Christopher Akiki, Ellie Pavlick, Suzana Ili{\'c},
  Daniel Hesslow, Roman Castagn{\'e}, Alexandra~Sasha Luccioni, Fran{\c{c}}ois
  Yvon, Matthias Gall{\'e}, et~al. 2022.
\newblock {BLOOM}: A 176b-parameter open-access multilingual language model.
\newblock \emph{arXiv}.

\bibitem[{Sutskever et~al.(2014)Sutskever, Vinyals, and
  Le}]{sutskever2014seq2seq}
Ilya Sutskever, Oriol Vinyals, and Quoc~V Le. 2014.
\newblock Sequence to sequence learning with neural networks.
\newblock \emph{NeurIPS}.

\bibitem[{Taori et~al.(2023)Taori, Gulrajani, Zhang, Dubois, Li, Guestrin,
  Liang, and Hashimoto}]{alpaca}
Rohan Taori, Ishaan Gulrajani, Tianyi Zhang, Yann Dubois, Xuechen Li, Carlos
  Guestrin, Percy Liang, and Tatsunori~B. Hashimoto. 2023.
\newblock Stanford alpaca: An instruction-following llama model.
\newblock \url{https://github.com/tatsu-lab/stanford_alpaca}.

\bibitem[{Touvron et~al.(2023)Touvron, Lavril, Izacard, Martinet, Lachaux,
  Lacroix, Rozi{\`e}re, Goyal, Hambro, Azhar et~al.}]{touvron2023llama}
Hugo Touvron, Thibaut Lavril, Gautier Izacard, Xavier Martinet, Marie-Anne
  Lachaux, Timoth{\'e}e Lacroix, Baptiste Rozi{\`e}re, Naman Goyal, Eric
  Hambro, Faisal Azhar, et~al. 2023.
\newblock Llama: Open and efficient foundation language models.
\newblock \emph{arXiv}.

\bibitem[{Vaswani et~al.(2017)Vaswani, Shazeer, Parmar, Uszkoreit, Jones,
  Gomez, Kaiser, and Polosukhin}]{vaswani2017attention}
Ashish Vaswani, Noam Shazeer, Niki Parmar, Jakob Uszkoreit, Llion Jones,
  Aidan~N Gomez, {\L}ukasz Kaiser, and Illia Polosukhin. 2017.
\newblock Attention is all you need.
\newblock \emph{NeurIPS}.

\bibitem[{Vilar et~al.(2022)Vilar, Freitag, Cherry, Luo, Ratnakar, and
  Foster}]{vilar2022prompting}
David Vilar, Markus Freitag, Colin Cherry, Jiaming Luo, Viresh Ratnakar, and
  George Foster. 2022.
\newblock Prompting {PaLM} for translation: Assessing strategies and
  performance.
\newblock \emph{arXiv}.

\bibitem[{Wang et~al.(2022)Wang, Li, Tan, Tu, Sun, and Liu}]{wang2022template}
Shuo Wang, Peng Li, Zhixing Tan, Zhaopeng Tu, Maosong Sun, and Yang Liu. 2022.
\newblock A template-based method for constrained neural machine translation.
\newblock In \emph{EMNLP}.

\bibitem[{Wei et~al.(2022)Wei, Bosma, Zhao, Guu, Yu, Lester, Du, Dai, and
  Le}]{wei2022finetuned}
Jason Wei, Maarten Bosma, Vincent Zhao, Kelvin Guu, Adams~Wei Yu, Brian Lester,
  Nan Du, Andrew~M Dai, and Quoc~V Le. 2022.
\newblock Finetuned language models are zero-shot learners.
\newblock In \emph{ICLR}.

\bibitem[{Wu et~al.(2023)Wu, Wang, Wan, Jiao, and Lyu}]{wu2023chatgpt4gec}
Haoran Wu, Wenxuan Wang, Yuxuan Wan, Wenxiang Jiao, and Michael Lyu. 2023.
\newblock {ChatGPT} or {Grammarly}? {E}valuating {ChatGPT} on grammatical error
  correction benchmark.
\newblock \emph{arXiv}.

\bibitem[{Xu et~al.(2023)Xu, Sun, Zheng, Geng, Zhao, Feng, Tao, and
  Jiang}]{xu2023wizardlm}
Can Xu, Qingfeng Sun, Kai Zheng, Xiubo Geng, Pu~Zhao, Jiazhan Feng, Chongyang
  Tao, and Daxin Jiang. 2023.
\newblock Wizardlm: Empowering large language models to follow complex
  instructions.
\newblock \emph{arXiv}.

\bibitem[{Yang et~al.(2023)Yang, Li, Zhang, Chen, and
  Cheng}]{Yang202ChatGPT4Summary}
Xianjun Yang, Yan Li, Xinlu Zhang, Haifeng Chen, and Wei Cheng. 2023.
\newblock Exploring the limits of {ChatGPT} for query or aspect-based text
  summarization.
\newblock \emph{arXiv}.

\bibitem[{Zhang et~al.(2023)Zhang, Haddow, and Birch}]{zhang2023prompting}
Biao Zhang, Barry Haddow, and Alexandra Birch. 2023.
\newblock Prompting large language model for machine translation: A case study.
\newblock \emph{arXiv}.

\end{thebibliography}
\bibliographystyle{acl_natbib}

\appendix

\section{Flores Full Sets}
\label{sec:flores-full}
We adopted the subset of Flores in order to directly compare with the results of ChatGPT and commercial MT systems from the previous evaluation report~\cite{jiao2023ischatgpt}, as stated in Section~\ref{sec:evaluation}. However, for a more convincing evaluation, we also test the LLMs on the full sets of Flores (i.e., with 1012 sentences), and list the results in Table~\ref{tab:mt-performance-floresfull}. Obviously, the trend of performance across systems still holds, and our "ParroT + Infer w/ No Err" setting performs the best.

\begin{table*}[t!]
\fontsize{10}{11}\selectfont
\centering
\caption{Translation performance of LLaMA models on Flores full sets.}
\begin{threeparttable}
\begin{tabular}{l rcrcrcrc}
\toprule
\multirow{2}{*}{\bf System} & \multicolumn{2}{c}{\bf De$\Rightarrow$En} & \multicolumn{2}{c}{\bf En$\Rightarrow$De} & \multicolumn{2}{c}{\bf Zh$\Rightarrow$En} & \multicolumn{2}{c}{\bf En$\Rightarrow$Zh} \\
\cmidrule(lr){2-3}\cmidrule(lr){4-5}\cmidrule(lr){6-7}\cmidrule(lr){8-9}
& BLEU & COMET & BLEU & COMET & BLEU & COMET & BLEU & COMET \\
\midrule
\multicolumn{9}{c}{\it Base Model: LLaMA-7b} \\
Vanilla & 3.0 & 58.2 & 1.7 & 48.8 & 2.7 & 52.0 & 0.1 & 48.6 \\
Alpaca & 36.3 & 87.6 & 23.0 & 81.1 & 16.4 & 81.3 & 8.3 & 56.5 \\
\hdashline
ParroT-T  & 37.4 & 87.9 & 26.9 & 83.7 & 19.7 & 82.8 & 26.1 & 79.2 \\
ParroT & 38.1 & 88.0 & 28.9 & 84.5 & 21.1 & 83.0 & 27.0 & 80.3 \\
~~~+ Infer w/ Prefer. & 34.6 & 87.5 & 24.8 & 83.9 & 18.1 & 82.4 & 24.2 & 79.0 \\
~~~+ Infer w/ No Err. & \bf 39.4 & \bf 88.4 & \bf 30.7 & \bf 84.6 & \bf 21.3  & \bf 83.4 & \bf 29.1 & \bf 81.2 \\
\bottomrule
\end{tabular}
\end{threeparttable}
\label{tab:mt-performance-floresfull}
\end{table*}

\section{General Tasks}
\label{sec:general-tasks}

In our experiments, we include the Alpaca multi-task dataset for finetuning ParroT to preserve its ability on general tasks.
We present two examples in Table~\ref{tab:general-abilities-case}, from question answering and code generation, respectively. The responses of both Alpaca and ParroT are correct, although the expressions may be different.
For comprehensive comparison, it is better to perform a quantitative evaluation and we leave it for future work.

\begin{table*}[t]
\fontsize{10}{11}\selectfont
\centering
\caption{Examples of general tasks. The base model is LLaMA-7b.}
\resizebox{0.99\textwidth}{!}{
\begin{tabular}{l p{14cm}}
\toprule
\bf System & \multicolumn{1}{c}{\bf Question Answering} \\
\midrule
& Tell me about alpacas. \\[1ex]
\hline
Alpaca & Alpacas are a species of camelid native to South America. They are domesticated and bred for their fiber, which is used in the production of knitted and woven items such as clothing, blankets, and rugs. Alpacas are typically kept as livestock, and can live up to 20 years in captivity. They are typically brown, white, or grey in color, and can grow up to 12 inches tall at the shoulder. \\[1ex]
\hdashline
ParroT & Alpacas are a species of domesticated South American camelids, bred primarily for their fiber. They are typically kept as pets, for their fleece, or for their meat. Alpacas are very social animals and live in herds, with a dominant male leader and a number of females and young. They are very intelligent and can be trained to respond to basic commands. Alpacas require a lot of space and good quality food in order to thrive. \\
\bottomrule
\bottomrule
\bf System & \multicolumn{1}{c}{\bf Code Generation} \\
\midrule
& Write a Python program that prints the first 10 Fibonacci numbers. \\[1ex]
\hline
Alpaca & numbers = [0, 1] \\
& for i in range(2, 10): \\
& \qquad numbers.append(numbers[i-2] + numbers[i-1]) \\[1ex]
& print(numbers)  \\[1ex]
\hdashline
ParroT & def fibonacci(n): \\
& \qquad if n == 0: \\
& \qquad\qquad return 0 \\
& \qquad elif n == 1: \\
& \qquad\qquad return 1 \\
& \qquad else: \\
& \qquad\qquad return fibonacci(n-1)+fibonacci(n-2) \\[1ex]
& print(fibonacci(10)) \\
\bottomrule
\end{tabular}
}
\label{tab:general-abilities-case}
\end{table*}

\end{document}